\pdfoutput=1

\documentclass[11pt]{article}

\usepackage[]{acl}

\usepackage{times}
\usepackage{latexsym}

\usepackage[T1]{fontenc}

\usepackage[utf8]{inputenc}

\usepackage{microtype}
\usepackage{enumitem}
\usepackage{multirow}
\usepackage{wrapfig}
\usepackage{colortbl}
\usepackage[normalem]{ulem}
\usepackage{microtype}
\usepackage{comment}
\usepackage[hang,flushmargin]{footmisc}

\setlist[itemize]{align=parleft,left=0pt,topsep=1mm,itemsep=0mm,parsep=1mm}
\definecolor{azure}{rgb}{0.0, 0.5, 1.0}
\definecolor{azure(colorwheel)}{rgb}{0.0, 0.5, 1.0}
\definecolor{nicegreen}{rgb}{0.0, 0.7, 0.1}
\definecolor{yw}{rgb}{0.01176, 0.5490, 0.5490}
\definecolor{ashblue}{rgb}{0.36, 0.54, 0.66}
\definecolor{ashgrey}{rgb}{0.7, 0.75, 0.71}
\definecolor{applegreen}{rgb}{0.55, 0.71, 0.0}
\definecolor{blue}{rgb}{0.0, 0.0, 1.0}
\definecolor{postechred}{rgb}{0.784, 0.003, 0.313}
\definecolor{ywg}{rgb}{0.9960, 0.8984, 0.5859}
\definecolor{ballblue}{rgb}{0.13, 0.67, 0.8}
\definecolor{cornellred}{rgb}{0.7, 0.11, 0.11}
\definecolor{darkcyan}{rgb}{0.0, 0.55, 0.55}
\definecolor{CuGray}{gray}{0.9}
\definecolor{airforceblue}{rgb}{0.36, 0.54, 0.66}
\definecolor{rev}{rgb}{0.784, 0.003, 0.313}
\definecolor{pink}{cmyk}{0, 0.7808, 0.4429, 0.1412}
\definecolor{amethyst}{rgb}{0.6, 0.4, 0.8}
\definecolor{black}{rgb}{0.0, 0.0, 0.0}
\definecolor{tb3_yellow}{rgb}{0.996, 1.0, 0.6}
\definecolor{tb3_orange}{rgb}{0.980, 0.8, 0.604}
\definecolor{tb3_red}{rgb}{0.972, 0.6, 0.6}
\definecolor{dimgray}{rgb}{0.41, 0.41, 0.41}
\definecolor{brickred}{rgb}{0.8, 0.25, 0.33}
\definecolor{bleudefrance}{rgb}{0.19, 0.55, 0.91}
\definecolor{blue(ncs)}{rgb}{0.265, 0.445, 0.765}
\definecolor{blue(ryb)}{rgb}{0.01, 0.28, 1.0}
\definecolor{orange}{rgb}{1.0, 0.49, 0.0}
\definecolor{Gray}{gray}{0.88}
\definecolor{green(ncs)}{rgb}{0.0, 0.62, 0.42}

\definecolor{kellygreen}{rgb}{0.3, 0.73, 0.09}

\newcolumntype{g}{>{\columncolor{CuGray}}c}
\newcolumntype{z}{>{\columncolor{CuGray}}l}

\renewcommand{\paragraph}[1]{\vspace{1mm}\noindent\textbf{#1.}\,\,}
\newcommand{\question}[1]{\vspace{1mm}\noindent\textbf{#1}\,\,}

\usepackage{xspace}

\makeatletter
\def\@fnsymbol#1{\ensuremath{\ifcase#1\or *\or \dagger\or \ddagger\or
   \mathsection\or \mathparagraph\or \|\or **\or \dagger\dagger
   \or \ddagger\ddagger \else\@ctrerr\fi}}
\makeatother

\def\onedot{.\@\xspace}
\def\eg{\emph{e.g}\onedot} 
\def\ie{\emph{i.e}\onedot}

\newcommand{\Fref}[1]{Fig.~\ref{#1}}
\newcommand{\Tref}[1]{Table~\ref{#1}}

\newcommand{\be}{\begin{eqnarray}}
\newcommand{\ee}{\end{eqnarray}}
\newcommand{\bee}{\begin{eqnarray*}}
\newcommand{\eee}{\end{eqnarray*}}

\newcommand{\matrixb}{\left[ \begin{array}}
\newcommand{\matrixe}{\end{array} \right]}

\usepackage{inconsolata}

\usepackage[utf8]{inputenc} 
\usepackage[T1]{fontenc}    
\usepackage{url}            
\usepackage{booktabs}       
\usepackage{amsfonts}       
\usepackage{nicefrac}       
\usepackage{microtype}      
\usepackage{xcolor}         
\usepackage{graphicx}
\usepackage{eucal}
\usepackage{subcaption}
\usepackage{multirow}
\usepackage{multicol}
\usepackage{amsmath}
\usepackage{amssymb}

\newcommand{\dataEmoji}{\includegraphics[height=.8em,trim=0 2em 0em 0]{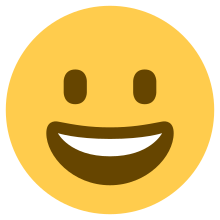}}
\newcommand{\datasetName}{\dataEmoji\xspace\textsc{SMILE}}

\title{\datasetName: Multimodal Dataset for Understanding Laughter in Video with Language Models}

\usepackage{mathrsfs}

\def\authorBlock{
    Lee Hyun${}^{1,2}\thanks{\ \ equally contributed}$ \ \thanks{\  \ work done at POSTECH} \qquad
    Kim Sung-Bin${}^{1  *}$ \qquad
    Seungju Han${}^{3}$ \qquad 
    Youngjae Yu${}^{4}$ \qquad
    Tae-Hyun Oh${}^{1,5,6}$ \vspace{3mm} \\ 
    ${}^{1}$Dept.~of Electrical Engineering and ${}^{5}$Grad.~School of Artificial Intelligence, POSTECH\\
    ${}^{2}$Samsung Advanced Institute of Technology \\
    ${}^{3}$Seoul National University \qquad ${}^{4}$Yonsei University\\ 
${}^{6}$Institute for Convergence Research and Education in Advanced Technology, Yonsei University\\
    {\tt \{hyunlee, sungbin, taehyun.oh\}@postech.ac.kr}
}
\author{\authorBlock}

\begin{document}
\maketitle

\begin{abstract}
Despite the recent advances of the artificial intelligence, 
building social intelligence remains a challenge.
Among social signals, laughter is one of the distinctive expressions that occurs during social interactions between humans.
In this work, we tackle a new challenge for machines to understand the rationale behind laughter in video, \emph{Video Laugh Reasoning}.
We introduce this new task to explain why people laugh in a particular video and a dataset for this task.
Our proposed dataset, \datasetName, comprises video clips and language descriptions of why people laugh. 
We propose a baseline by leveraging the reasoning capacity of large language models (LLMs) with textual video representation. Experiments show that our baseline can generate plausible explanations for laughter. We further investigate the scalability of our baseline by probing other video understanding tasks and in-the-wild videos.
We release our dataset, code, and model checkpoints on \href{https://github.com/postech-ami/SMILE-Dataset}{https://github.com/postech-ami/SMILE-Dataset}.
\end{abstract}

\section{Introduction}
\begin{flushright}
\textit{``Laughter is the shortest distance between two people.''} \\
---\textsc{Victor Borge}
\end{flushright}

We, human beings, are 
immersed in laughter.
Laughter is a distinctive non-verbal social signal, associated with bonding, agreement, affection, and emotional regulation~\citep{scott2014social}.
It is often purposedly elicited to establish intimacy~\cite{stauffer1999let}, grab attention~\cite{wanzer2010explanation}, or build faith~\cite{vartabedian1993humor}; \ie, serving as a powerful medium to express a wide range of social and emotional implications beyond the capacity of mere words. 
Thus, understanding laughter is a crucial problem with huge potential in artificial social intelligence~\cite{bainbridge1994artificial,williams2022supporting,dautenhahn2007socially} to build empathetic machines with human-machine interaction~\cite{lee2017bayesian,nijholt2017humor,inoue2022can}.
However, understanding and modeling laughter reactions is challenging.
Even a simple joke is associated with language skills, context knowledge, theory-of-mind, abstract thinking, and social perception, and 
complex entanglement of these makes laughter reaction arguably the most complex cognitive attribute humankind may have \cite{mcdonald2013philosophy}.
\begin{figure*}[t]
    \centering
    \includegraphics[width=\textwidth]{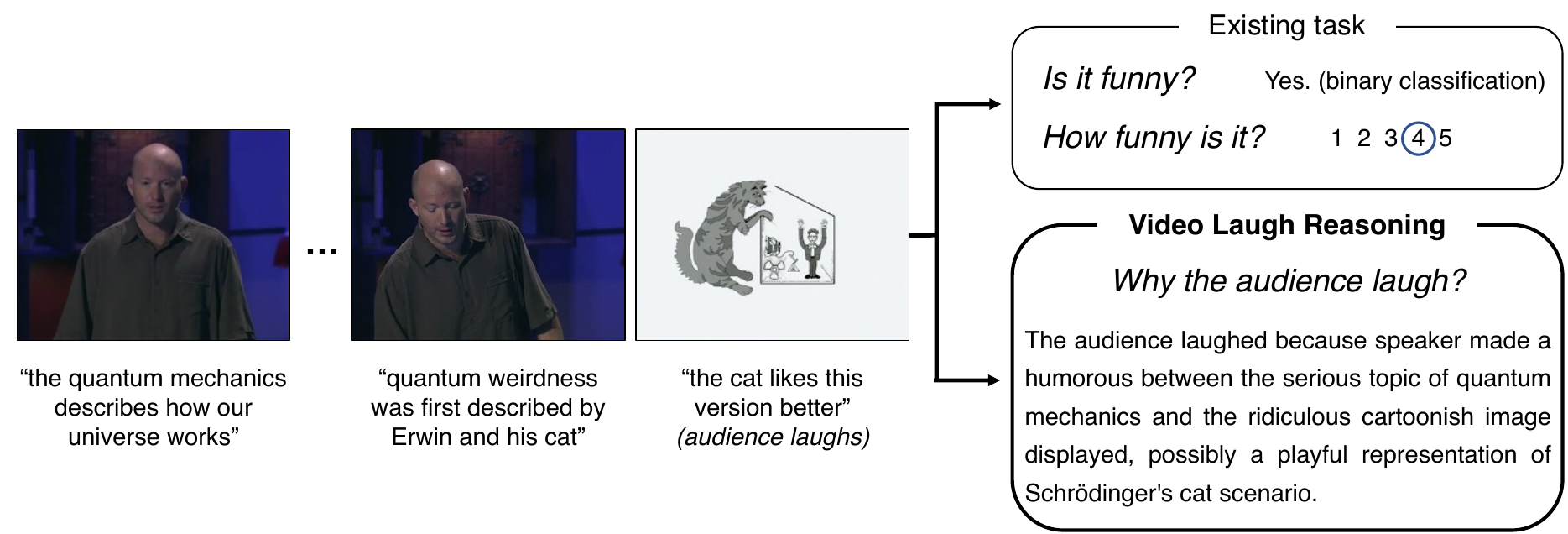}
    \caption{Why do people laugh? We present \textit{Video Laugh Reasoning}, a new task to interpret the reasons behind laughter in a video.}
    \label{fig:teaser}
    
\end{figure*}

In this work, we take the first 
stepping stone to tackle the challenge of understanding laughter by introducing a task, \emph{Video Laugh Reasoning} that aims to interpret the reasons behind laughter in a video.
For this task, we curate a new dataset, \datasetName, consisting of video clips and corresponding text annotations explaining reasons for laughter.
We probe through the question ``Why do people laugh?'' and reason through the answer in a language form; thus, we define the task as a free-form text generation task in which the model generates an explanation for the laughter with a given video clip (See Figure~\ref{fig:teaser}).


While reasoning laughter by answering the question is an effective way of probing the level of understanding, 
laughter itself has an inherently complex nature which can be influenced by 
diverse factors~\cite{apte1985humor, provine2001laughter, martin2003individual, martin2018psychology}, \eg, the subjectivity~\cite{warren2021makes}, context knowledge~\cite{nijholt2017humor}, and multimodality~\cite{hasan-etal-2019-ur}. 
To build a clearer resource for understanding laughter and its social norm behind it, we design the dataset to focus on \emph{audience laughter}, a cohesive form from social influence in distinct contexts~\cite{greatbatch2003displaying}, and thereby alleviating the subjectivity associated with individual laughter. 
Also, for our task, we propose a baseline based on large language models (LLMs) with \emph{multimodal textual representation} by converting multimodal attributes and features on video into a textual format.

Our experimental results show that the proposed baseline, incorporating LLM's reasoning capability with multimodal textual representation, can generate plausible explanations of the reason for laughter. 
Our data analysis and ablation study reveals that multimodal information plays a role in understanding laughter. 
We further explore the scalability of utilizing LLM with textual representation by applying it to other video understanding tasks and in-the-wild videos.

Our major contributions are threefold: 1) 
proposing \emph{Video Laugh Reasoning}, a new task for understanding the reason behind laughter in a video, 2) building \datasetName,
a new dataset that comprises video and explanation for laughter reason,
and 3) 
 presenting a baseline using LLM with multimodal textual representation for laugh reasoning task and its scalability.
\section{Related Work}
\paragraph{Understanding laughter} 
Laughter plays a key role in social interactions, such as bonding, agreement, affection, and emotional regulation~\cite{scott2014social}. 
Given its importance in social interactions, seminar works tackle to detect laugh-inducing moments, specifically focusing on humor or sarcasm. 
Several methods~\cite{annamoradnejad2020colbert, weller-seppi-2020-rjokes} rely primarily on transcripts for humor detection. 
As laughter occurs with multimodal information, such as variations in tone or facial cues, there are attempts to incorporate audio and text cues from videos~\cite{bertero-fung-2016-deep, alnajjar-etal-2022-laugh}, or even include visual cues~\cite{castro-etal-2019-towards, hasan-etal-2019-ur, ray-etal-2022-multimodal} to pinpoint the occurrences of humor. 
Yet they focus on detecting whether a certain situation induces laughter or predicting the intensity of laughter, without providing explanations for the underlying reasons behind the laughter (See Figure~\ref{fig:teaser}).
Moreover, despite the availability of datasets for understanding the types and characteristics of laughing moments~\cite{urbain-etal-2010-avlaughtercycle, mckeown2012ilhaire, dupont2016laughter}, no dedicated dataset is available for comprehending the context surrounding laughter.
Few works~\cite{PaLM,hessel-etal-2023-androids, ko2023can} have attempted to reason about laughter or jokes.
However, their scope differs from ours, as they focus on providing instant textual descriptions of humor or cartoon images accompanied by text.
To the best of our knowledge, we are the first to introduce the task of understanding the reason for laughter within videos, accompanied by 
our comprehensive dataset. 

\paragraph{Multimodal reasoning}
Multimodal reasoning is a complex task
aiming to equip machines with the capability to parse, analyze, and logically reason about the given multimodal context. A widely explored reasoning task is a question answering (QA) on images~\cite{antol2015vqa,gao2015you,zhu2016visual7w} or video~\cite{lei-etal-2018-tvqa,tapaswi2016movieqa}, which requires understanding the question, referencing the appropriate context, and selecting the correct answer.
Similarly, commonsense reasoning~\cite{vedantam2015learning,yatskar-etal-2016-stating,wu2016ask} is another type of reasoning, demanding a more profound level of understanding and the ability to infer unstated information. Our task includes commonsense reasoning in that laughter is often elicited by exploiting external contexts, rather than merely understanding underlying phenomena.

Several methods~\cite{zellers2019recognition,vicol2018moviegraphs,zadeh2019social} have attempted to learn and reason about the social interactions in the video.
For instance, Visual Commonsense Reasoning (VCR)~\cite{zellers2019recognition} unifies reasoning about diverse commonsense phenomena, while Social IQ~\cite{zadeh2019social} aims to teach social intelligence by providing a broad range of social and behavioral situations to a machine. However, these approaches give less attention to a deeper understanding of laughter itself---a complex non-verbal signal integral to social interactions. Unlike the prior arts, we specifically focus on the task of reasoning human laughter. We posit this as a significant stride towards understanding important social signals frequently encountered in daily life, thus contributing a new perspective to multimodal reasoning and understanding tasks.

\paragraph{Models for multimodal reasoning}
To tackle multimodal reasoning, one approach is to design pretraining methods~\cite{lu2019vilbert, li2019visualbert} that learn the joint vision and language representations. More recently, the combination of large-scale vision and language models (VLM) has demonstrated remarkable performance in multimodal reasonings~\cite{2023videochat, UnifiedIO, damonlpsg2023videollama,OFA, han2023champagne}.

An alternative approach for multimodal reasoning utilizes text as a unified representation and large language models (LLM) with minimal or without training. For instance, Socratic Model~\cite{socratic} employs language to combine complementary knowledge from various pre-trained models for tackling a wide range of tasks. Similarly, \citet{wang2022language} converts the visual attributes into the text representation to prompt a frozen LLM for diverse video-language tasks. In this work, we conduct extensive experiments on our proposed laugh reasoning task and show the effectiveness of using text as an intermediate representation.
\begin{figure*}
    \centering
    \includegraphics[width=\textwidth]{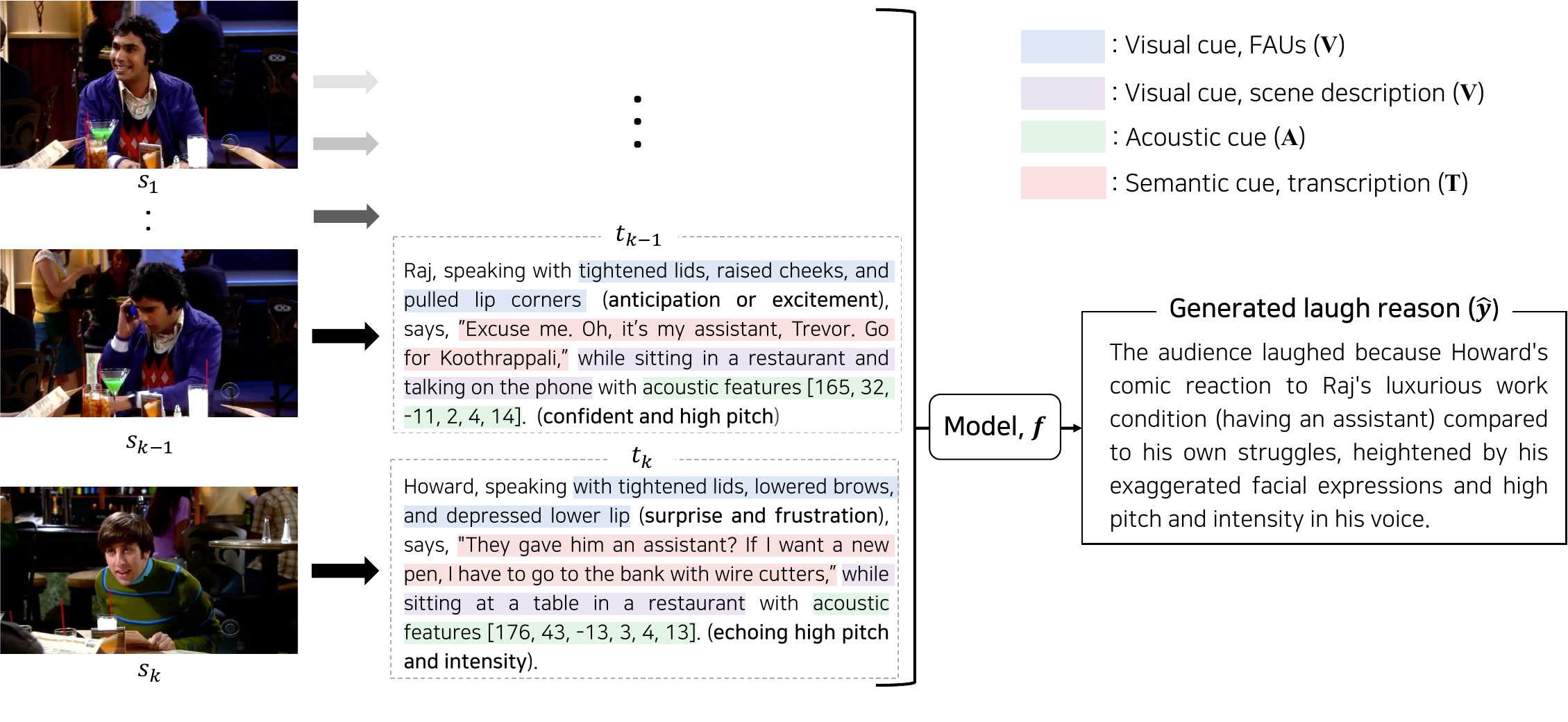}
    \caption{\textbf{Video Laugh Reasoning task and multimodal textual representation.} 
    Each video clip ($v$) is trimmed into list of video segments ($s_i$), and each video segment is encoded into textual representation ($t_i$).
    The textual video representation consists of visual cues ($V$), acoustic cues from speech ($A$), and semantic cue (transcript, denoted as $T$). Then, we use LLM to generate why the audience laughs at the given video with the prompt. 
    The bold text in parentheses on the $t$ shows that LLM is semantically aware of the textual video representation.}
    \label{fig:task_definition}
\end{figure*}

\section{Task Definition and Dataset}

In this section, we introduce our \emph{Video Laugh Reasoning} task and our dataset for it.


\vspace{-1mm}

\subsection{Task Definition and Baseline}
We present \emph{Video Laugh Reasoning}, a task that challenges the model to understand reasons for laughter in a given video. We pose our task as a generation problem, enabling the model to explain why a particular situation incited laughter in the video. 
We 
define this task as, $\hat{y} = f(v)$, where $\hat{y}$, $f$, and $v$ stand for the generated explanation about laughter reason, the model, and the given video clip.

For this task, we propose a baseline that utilizes the reasoning capacity of LLM. To ensure compatibility of input $v$ with the language model, we convert videos into \emph{multimodal textual representation} that preserve multimodal information from video, such as visual, acoustic, and semantic cues. We compose visual cues with facial expressions\footnote{We use facial action units~\cite{ekman1978facial}.} and scene descriptions\footnote{We use video captioning model~\cite{wang2022internvideo}.} to perceive human-specific and scene-wide contextual information. For acoustic cues, we extract the mean and the variance of pitch, intensity, jitter, and shimmer from speech to capture. 
We simply use transcripts of the speech from the videos for semantic cues (See Figure~\ref{fig:task_definition}).

Using textual representation as input and LLM as model $f$, we can rewrite the task formula as, $\hat{y} = f(\mathcal{P}, \ \left\{ t_1,\  t_2, ..., \  t_k \right\})$, where $\mathcal{P}$ stand for the prompt that describes input representation and instructing the laugh reasoning task to language models and $t$ is multimodal textual representation converted from the given video clip $v$. See Appendix~\ref{app:A} for details about how to convert video into textual representation.

\subsection{Dataset}
\paragraph{Data collection}
We present \datasetName, a curated dataset encompassing 887 video clips, each paired with a language description about the reason for laughter for the corresponding video clip. This pairing facilitates supervised training for the laugh reasoning task. The dataset focuses on audience laughter among many types of laughter since audience laughter usually has a clearer signal than other laughter and represents a general and cohesive form of laughter. To encompass a wider range of videos that contain situations where audiences laugh, we construct our dataset using two different sources: \emph{TED talks and sitcoms}.\footnote{We source the video clips from MUStARD~\cite{castro-etal-2019-towards} and UR-Funny dataset~\cite{hasan-etal-2019-ur}.} 


We curate video clips that span between 10 and 90 seconds for \emph{TED talks} and 7 and 60 seconds for \emph{sitcoms}. If a video is too short, it might fail to provide sufficient context for laughter. In contrast, if a video is too long, it may dilute specific laughter-inducing contexts with unrelated 
information. The average duration for TED talk clips is 
longer than \emph{sitcoms}, given the protracted nature of talks.

Given that a single video clip often contains multiple instances of laughter, we focus on the last laugh in a clip for easier annotation. We only use video clips that meet the following filtering criteria, using a laugh detector~\cite{gillick2021robust} to identify audience laughter instances.
Our filtering criteria are: 
laughter should 
last at least 0.5 seconds, and 
be no more than 1 second interval 
between the video clip's last utterance and the onset of laughter.
The latter criterion filters out the laughter events that are not related to the punchlines but are induced by something else.
After this pre-processing, our final dataset comprises 484 sitcom and 403 TED talk video clips. Table~\ref{tab:dataset_statistics} shows the statistics of our dataset.
\begin{figure*}
    \centering
    \includegraphics[width=1\textwidth]{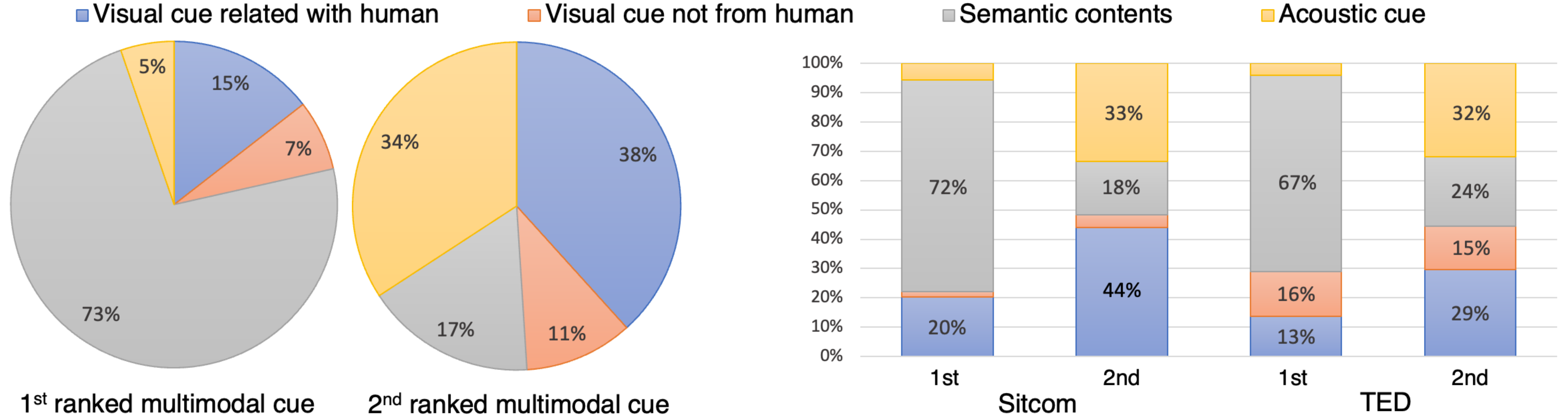}
    \caption{\textbf{Which multimodal cue is important to reason the laughter?}
    While semantic content is the most influential in causing laughter, the $2^{\text{nd}}$ ranked modality cues are diverse, suggesting that multiple modality information can simultaneously influence laughter.}
    \label{fig:data_analysis_multi}
    
\end{figure*}

\begin{table}
  \centering
  \resizebox{0.95\linewidth}{!}{
    \begin{tabular}{l|c}
    \toprule
    Number of Video Clips & 887 \\
    Number of Train/Val/Test & 727 / 80 / 80 \\
    Number of Video Segments & 4,434 \\
    Avg. number of Segments per clip ($k$) & 4.4 \\
    Avg. duration of Video Clips & 27.5 sec. \\
    Avg. duration of Video Segments & 6.2 sec. \\
    \bottomrule
    \end{tabular}  
    }
    \caption{\textbf{Statistics of our dataset.} We split our dataset into train, validation, and test sets with the ratio of 8:1:1. Avg. denotes average.}
  \label{tab:dataset_statistics}
  
\end{table}

\begin{figure*}[t]
    \centering
    \includegraphics[width=\textwidth]{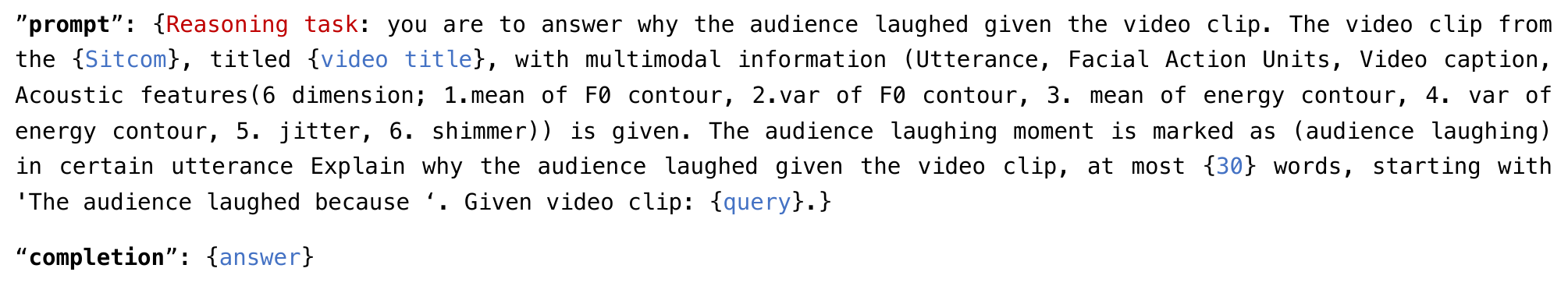}
    \caption{\textbf{Prompt for laugh reasoning experiments on GPT3.} 
    The prompt is fed into GPT3~\cite{GPT3} for fine-tuning, zero-shot learning, and in-context learning. For in-context learning, three random samples of prompt-answer pairs from the training set are given to GPT3.
    We manually change video types (sitcom or TED) and video title using the meta information of video clips. The query stands for multimodal textual representation $m$ of the video clip. The length of the generated output is also variable, with a maximum of 30 words for sitcoms and 40 words for TED talks, considering each video type's characteristics.}
    \label{supp_fig:prompt_reason}
\end{figure*}

\paragraph{Annotation for laughter reason} \label{sec:32}
We employ human annotators from Amazon Mechanical Turk (AMT) to label videos with reasons for laughter.
Given the inherently subjective nature of humor and the extensive variability in laughter triggers, constructing ground truth (GT) by free-form annotation is challenging.
To mitigate these issues, we utilize the language model
to generate candidates for laughter reasons, these candidates are subsequently presented to annotators with the corresponding video clip to choose the most appropriate explanation among them and refine it. If none of the candidates were suitable, we instruct them to write the reason in a free form. 

After annotation, 
we verify
all GT and manually refine it if it is not plausible for laughter reasons with video. This approach reduces 
the annotation
workload by interacting LLM and humans, developing a more concise GT for this complex and subjective task. 
Finally, our dataset is formed 
as $\mathcal{D} = \left\{v, \ y \right\}$, where $y$ is a GT explanation for laughter in the video clip $v$. See Appendix~\ref{app:B} for details about the human annotation process and the post-processing. Also, refer to Appendix~\ref{app:F_AMT} for the details about the AMT configuration.

\subsection{Data Analysis}
\paragraph{Which multimodal cue is important to infer the reason for laughter}\label{sec34}
We conduct a human evaluation to understand our dataset better. The annotators are requested to rank the multimodal cues in perspective of which cues are related
to laughter in the video. The rank annotation provides insight into which modality information is crucial for the cause of the laughter for each case.

For each video clip, we present annotators four choices: 1) visual cues from human; \textit{e.g.}, facial expression and body gesture, 2) visual cues not from human; \textit{e.g.}, backgrounds or images and props, 3) semantic contents; \textit{i.e.}, transcription, and 4) acoustic cues; \textit{e.g.}, speech tone or intensity. We ask them to choose two modality cues that are the most relevant for inducing laughter. The pie chart on the left in Figure~\ref{fig:data_analysis_multi} shows the modality importance statistics for our dataset. While the reason for laughter is primarily driven by semantic contents, the second most effective cue varies across different modalities, indicating that the various modalities in the video contribute to the reason for laughter.
\begin{table}
  \vspace{2mm}
  \centering
    \resizebox{0.49\textwidth}{!}{\begin{tabular}{lccccc}
    \toprule
    Model & BLEU$_4$ ($\uparrow$) & METEOR ($\uparrow$) & ROUGE$_L$ ($\uparrow$) & BERTScore ($\uparrow$) & Win rate \\
    \midrule						
    Video model & 0.226 & 0.236 & 0.398 & 0.427 & 24\% \\
    LLM + multimodal & \textbf{0.270} & \textbf{0.256} & \textbf{0.432} & \textbf{0.496} & \textbf{76\%} \\	
    \bottomrule
\end{tabular}}\\

  \caption{\textbf{Comparison with video model.} We compare the video model trained on raw video and transcripts with LLM trained on multimodal textual representation. 
  We use Video-LLaMA~\cite{damonlpsg2023videollama} and LLaMA~\cite{llama} for video model and LLM, respectively.}    
  \label{tab:baseline}
  
\end{table}
The bar chart on the right in Figure~\ref{fig:data_analysis_multi} shows the elements that induce laughter in two video types of our dataset.
Notably, visual cues unrelated to humans, such as backgrounds or images, significantly trigger more laughter in TED than in sitcoms.
TED videos often exhibit the speaker's presentation slides, making non-human visual cues more influential for eliciting laughter.
Conversely, visual cues such as facial expressions and body gestures have a higher probability of causing laughter in sitcoms than in TED. 
This difference is because sitcoms mainly center around the characters' dialogues, so visual cues from human actors are more crucial.  
See Appendix~\ref{app:C_data_analysis} for additional data analysis.

\section{Experiment} \label{sec:exp}
\begin{table*}[t]
  \vspace{2mm}
  
  \centering
  \resizebox{1.0\textwidth}{!}{\begin{tabular}{cccccccc}
    \toprule
    Model & Num. of parameters & Modality & BLEU$_4$ ($\uparrow$) & METEOR ($\uparrow$) & ROUGE$_L$ ($\uparrow$) & BERTScore (F1) ($\uparrow$)\\
    \midrule						
    \multirow{2}{*}{LLaMA (FT)}& \multirow{2}{*}{13B} & T  & 0.250 & 0.245 & 0.432 & 0.493\\
     && A+V+T  & 0.270 & 0.256 & 0.453&0.496\\ 
     \midrule
     \multirow{2}{*}{GPT-3 (zero-shot)} & 
    \multirow{2}{*}{175B} & T  & 0.126 & 0.155& 0.313 & 0.389\\
     && A+V+T  & 0.157 & 0.184 & 0.364&0.454\\
     \multirow{2}{*}{GPT-3 (3-shot)} & \multirow{2}{*}{175B} & T  & 0.187 & 0.198 & 0.368 & 0.431\\
     && A+V+T  & 0.232 & 0.230 & 0.413&0.476\\
\multirow{2}{*}{GPT-3 (FT)} & \multirow{2}{*}{175B} & T  & 0.230 & 0.243 & 0.429 & 0.488\\
     && A+V+T  & \textbf{0.279} & \textbf{0.267} & \textbf{0.475} &\textbf{0.523}\\
    \bottomrule
    
  \end{tabular}}\\
  \caption{\textbf{Evaluation on laugh reasoning with LLMs.} We evaluate whether the model can explain why the audience laughed. {We fine-tune two LLMs, GPT-3~\cite{GPT3} and LLaMA~\cite{llama} on our dataset, SMILE. We use GPT-3 for in-context (3 shots) and zero-shot experiments. Each modality cue in our dataset is denoted as Transcript (T), Audio (A), and Visual (V). FT denotes fine-tuning the model.}}
  \label{tab:results}
\end{table*}

We split our dataset into 5 cross-validation splits except for the test set. 
We fine-tune two LLMs, GPT-3~\cite{GPT3} and LLaMA~\cite{llama} with the training set and use the test set for evaluation.

\paragraph{Implementation details} We use the official GPT-3~\cite{GPT3}, a non-free commercial version, as follows.
We utilize the \emph{davinci-text-002} model of GPT-3~\cite{GPT3} for the zero-shot and in-context learning experiments. Examples of the prompts for both tasks are shown in Figure~\ref{supp_fig:prompt_reason}. The ``prompt'' provides the context of the task and the multimodal cues of the video, and ``completion'' provides the reason for the laughter. 
The zero-shot setup only takes ``prompt'' and generates the reason for the laughter, while the in-context learning setup is given with additional three randomly labeled samples from the training set as few-shot examples. More implementation details including LLaMA are in Appendix~\ref{app:D_implementation}.

\paragraph{Evaluation metrics}
We utilize both quantitative metrics and human evaluation.
We use metrics commonly employed for evaluating language generation tasks, including BLEU$_4$~\cite{papineni-etal-2002-bleu}, METEOR~\cite{banerjee-lavie-2005-meteor}, ROUGE$_L$~\cite{lin-2004-rouge}, and BERTscore~\cite{zhangbertscore}.
For the human evaluation, we gather assessments from 3 crowd-workers per test sample by asking them to select their preferred explanation for laughter from a pair of options and take a majority vote to determine a winner. We calculate the average win rate (\%) over the test set.

\subsection{Comparison with video model}
In addressing the laugh reasoning task, a direct
method is to train a video model with raw video input. We compare the video model with our baseline, which utilizes LLM with multimodal textual representation.
We fine-tune each model and conduct the quantitative and human evaluations (win rate), as shown in Table~\ref{tab:baseline}. The LLM-based baseline outperforms all metrics, indicating that 
our multimodal 
textual representation incorporates LLM's capacity to understand the reason for laughter in the video.

\begin{table}
\centering
\resizebox{1.0\linewidth}{!}{
  \begin{tabular}{lll|cc} 
    \toprule
     & A & B & A wins (\%) & Fleiss'-$\kappa$ \\
    \cmidrule{1-5}
    Q1 & GPT-3 (A+V+T) & GPT-3 (T)  & 72.2 & 0.43\\
    Q2 & GPT-3 (FT) & GPT-3 (3-shot)  & 77.8 & 0.31 \\
    Q3 &GPT-3 (FT) & LLaMA (FT)  &  56.6 & 0.49 \\
    Q4 &Human & GPT-3 (FT) & 66.2 & 0.42 \\

    \bottomrule
  \end{tabular}
}
\caption{\textbf{Pairwise human evaluation.} Except for Q1, we use all modality (A+V+T) for training. We use Fleiss'-$\kappa$~\cite{fleiss2013statistical} for assessing the reliability of agreement. Q1-Q4 denote corresponding evaluation in \S~\ref{sec:42_eval}.}
\vspace{-4mm}
\label{tab:human_winrate}  
\end{table}
\subsection{Evaluation} \label{sec:42_eval}
\begin{figure*}[t]
    \centering
    \includegraphics[width=\textwidth]{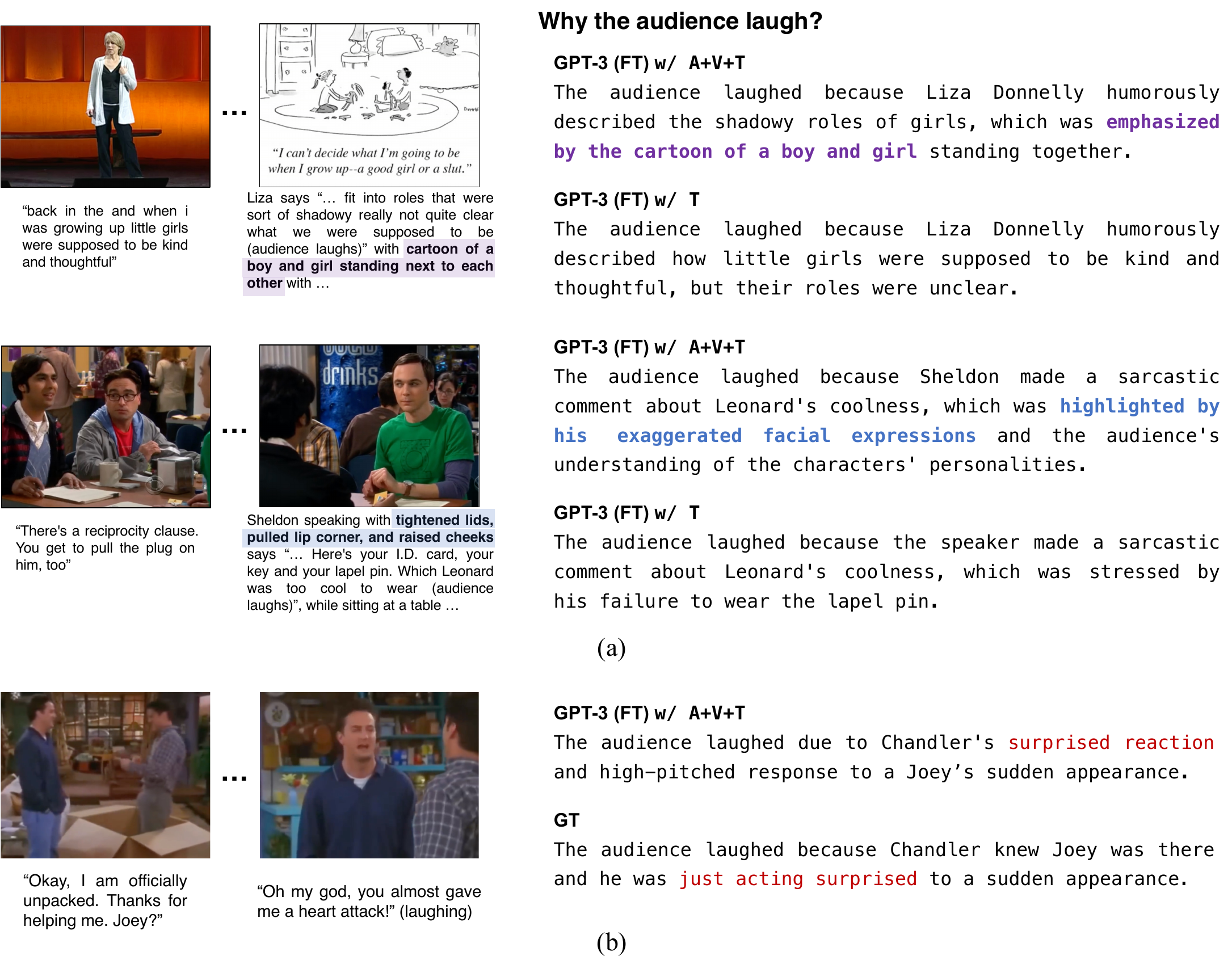}
    
    \caption{\textbf{Qualitative results on laugh reasoning.} For the examples in (a), GPT-3~\cite{GPT3} fine-tuned on our dataset (denoted FT w/ A+V+T) understands the reasons for laughter by referencing multimodal cues. In contrast, the model fine-tuned using the transcript-only (denoted FT w/ T) manages to understand the reasons partially. The visual cues (scene description) are crucial for capturing ``joey's sudden appearance'' which is important to infer the reason for laughter in (b).
    }
    \label{fig:qual}
    
\end{figure*}

We analyze our baseline on laugh reasoning in various setups. 
We utilize both quantitative and human evaluation. 
Quantitative results are in Table~\ref{tab:results}, and the results of human agreements are in Table~\ref{tab:human_winrate}. Our evaluations aim to address four key questions.

\question{Q1. Does multimodal information help for laugh reasoning?}
Yes, incorporating all modality cues for training enhances the performance of the laughter reasoning task compared to using transcripts alone (Table~\ref{tab:results}). The model trained with all modalities preferred in 72.2\% of the test set compared to the transcript-only model as shown in Table~\ref{tab:human_winrate}. Furthermore, \Fref{fig:qual} (a) supports this, showing that the model trained with all modalities can effectively distinguish the reasons for laughter by utilizing multimodal information, whereas a transcript-only model only achieves a partial understanding.

\question{Q2. Does the fine-tuning step help for a laugh reasoning?}
Yes, fine-tuned models outperform zero-shot/in-context models in both quantitative evaluation and human preference. It 
shows that our dataset nicely infuses the video laugh reasoning capacity to LLM.

\question{Q3. Do bigger models generate better reasons for laughter?}
Yes, GPT-3 (175B) surpasses LLaMA (13B) in both quantitative evaluation and human preference, as shown in Table~\ref{tab:results} and \ref{tab:human_winrate}.

\question{Q4. Does the model explain the reason for laughter as well as humans?}
No, the human-annotated laughter reasons are preferred by 66.2\% than those generated by fine-tuned GPT-3 (our best model) as shown in Q4 of Table~\ref{tab:human_winrate}. Figure~\ref{fig:qual} (b) provides an example illustrating the comparison between human-annotated reasons (GT) and generated reason for laughter. In this sample, all crowd workers prefer GT because the model struggles to distinguish the subtle difference between surprise and posed surprise, while the human-annotated reason successfully captures it. 




\begin{table}[t]
\centering
\resizebox{1.0\linewidth}{!}{
  \begin{tabular}{l|cc} 
    \toprule
    \multirow{2}{*}{Model} & MUStARD & UR-FUNNY \\
            & Acc. (\%) ($\uparrow$) & Acc. (\%) ($\uparrow$) \\
    \cmidrule{1-3}
    TFN~\cite{zadeh-etal-2017-tensor} & 68.6  & 64.7 \\
    CMFN~\cite{hasan-etal-2019-ur} & 70.0  & 65.2 \\
    MISA~\cite{hazarika2020misa} & 66.1  &  70.6 \\
    BBFN~\cite{han2021bi} & 71.4 & 71.7 \\
    MUStARD++~~\cite{ray-etal-2022-multimodal} & 74.2 & - \\
    MAG-XLNet~\cite{rahman-etal-2020-integrating} & 74.7 & 72.4 \\
    MuLoT~\cite{pramanick2022multimodal} & 76.8 & 73.9 \\
    \cmidrule{1-3}
    Ours (w/ LLaMA) & 77.5  & 75.1  \\
    Ours (w/ GPT-3) & \textbf{79.0}  & \textbf{77.9}  \\
    \bottomrule
  \end{tabular}
}
\caption{\textbf{Evaluation results of the humor \& sarcasm detection task.} All models use text, visual, and acoustic information from videos for training.}
\vspace{-3mm}
\label{tab:detection}

\end{table}


\begin{figure*}[t]
    \centering
    \includegraphics[width=1.0\textwidth]{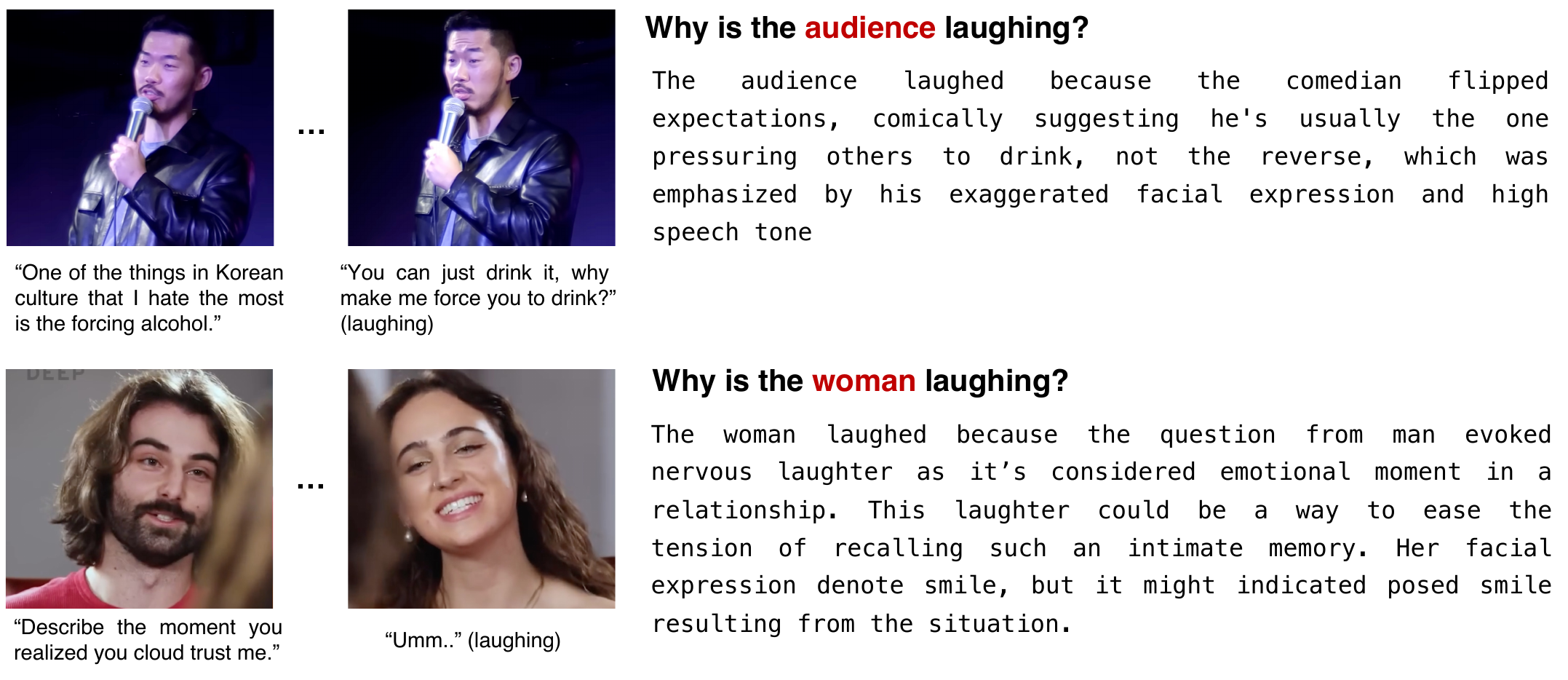}
    \vspace{-4mm}
    \caption{\textbf{Examples of in-the-wild videos.} We try to generate explanations for the laughter in the videos from standup comedy and intimate conversation. The results show that we can generate valid explanations for laughter in other videos.}
    \label{fig:wild}
    
\end{figure*}
In summary, for the laugh reasoning task, multimodal information, a large model, and infusing reasoning capacity with our dataset are important. While the trained model does not surpass human capabilities, the use of LLM with multimodal textual representation enables us to generate plausible explanations about the reason for laughter in videos. See Appendix~\ref{app:E_addi_exp} for additional experiments.

\section{Discussion}
In this section, we discuss the scalability of utilizing large language models with textual video representation by conducting evaluations on other tasks and on in-the-wild videos.

\subsection{Evaluation on other tasks}
Apart from laugh reasoning, we conduct humor detection and sarcasm detection tasks, which classify given video contains humor (sarcasm) or not (i.e., binary classification). We use UR-FUNNY~\cite{hasan-etal-2019-ur} and MUStARD~\cite{castro-etal-2019-towards}, which are representative benchmarks for these tasks. We cast the original binary classification problem as a text generation problem to integrate into our system. Formally, we can define the task as, $\hat{b} = f(\mathcal{P},\   \left\{ t_1,\  t_2, ..., \  t_k \right\})$, where $\hat{b}$ denote predicted binary class in text format ("Yes" or "No"), and $\mathcal{P}$ is prompt for instructing LLMs about the task and input representation.

We follow the same train/test split, and evaluation procedure as in the benchmark for measuring the accuracy of each detection task. We use LLaMA and GPT-3 for training with textual representation converted from the video in the training set of each benchmark dataset. Table~\ref{tab:detection} shows that our method achieves strong performance\footnote{We do not compare with FunnyNet~\cite{liu2022funnynet} as they use an additional large-scale dataset for training.} on both tasks. 
This experiment highlights the scalability of utilizing LLMs with textual representation in various video understanding tasks.

\subsection{Evaluation on the in-the-wild videos} \label{5.2}
We extend our laughter reasoning to in-the-wild videos, encompassing different video types and laughter contexts compared to our dataset. 
First, we evaluate our approach on a video clip from a stand-up comedy, which has similar audience laughter patterns to those in our dataset. We convert the video into a textual representation and infer the reason for the audience laughing. Figure~\ref{fig:wild} shows that the model can generate a plausible explanation for the reason for laughter in stand-up comedy. 

Next, we test on a video clip featuring an intimate conversation between a married couple. In this case, the laughter originates from the speakers themselves rather than from the audience. 
As this does not belong to the comedic genre but rather a sincere conversation between two people, it is more likely that non humor-based laughter, such as nervous or social laughter, may occur. Figure~\ref{fig:wild} shows that the model can also understand the nervous laughter used to alleviate tension or awkwardness in the situation.

\section{Conclusion}
In this paper, we aim to understand the reason behind laughter by introducing~\emph{Laugh Reasoning} task, accompanied with 
SMILE dataset.
While the model did not surpass human capabilities, we show that the model can generate plausible explanations about laughter reason, underlining that multimodal cues in our dataset nicely infuse the laugh reasoning capacity to the model. We also show the results applied to other tasks and other types of video, hinting at the scalability of utilizing LLM with multimodal textual representation.

\paragraph{Limitation \& future direction}
Our LLM-based baseline serves as the initial method for laugh reasoning task and has a margin to improve. For the multimodal textual representation, as it is a primitive form for capturing human social interaction in the video, we can enhance it with diverse attributes such as gesture, eye gaze, and relationship or use other representations such as scene graph. 
Our work mainly focuses on audience laughter as the first stepping stone toward understanding laughter due to its distinct and cohesive signal, while there are diverse mechanisms behind laughter. Recognizing this, enriching our work with diverse video types like vlogs, movies, and talk shows is a promising direction to capture a broader range of laughter, as we show the possibility in \S~\ref{5.2}. 

\paragraph{Potential application \& broader impact} 
Our work can be regarded as a stepping stone toward developing socially intelligent agents that understand
and appropriately create non-verbal cues, such as laughter, playing a crucial role in building rapport,
expressing emotions, and creating deep emotional exchanges~\cite{tickle1990nature, argyle1972non}. Such advancement moves us beyond the capabilities of current dialogue agents, e.g., ChatGPT or Alexa, which mostly focus on verbal signals. Incorporating 3D talking head methods~\cite{Sung-Bin_2024_WACV, zhao2024media2face} could offer the way agents are visualized, enabling more expressive and multimodal interactions with users.

\paragraph{Acknowledgement}
This work was supported by Institute of Information \& communications Technology Planning \& Evaluation (IITP) grant funded by the Korea government (No.2022-0-00290, Visual Intelligence for Space-Time Understanding and Generation based on Multi-layered Visual Common Sense and No.2022-0-00124, Development of Artificial Intelligence Technology for Self-Improving Competency-Aware Learning Capabilities and No.2021-0-02068, Artificial Intelligence Innovation Hub) and NCSOFT.

\clearpage
\bibliography{custom}
\clearpage
\appendix

\section{Multimodal Textaul Representation} \label{app:A}
In this section, we explain how to convert video into multimodal textual representation.
Videos are multimodal, which include visual, acoustic, and semantic cues (i.e., transcription). We encode video clips into textual representation, embracing their multimodal information, so that we can leverage the pre-trained knowledge of LLMs while exploiting multimodal inputs in our baselines. First, starting with a video clip, we build a list of video segments by trimming the clip based on the utterances. The definition of the utterance varies upon to the source of the video: for TED talks, each sentence is defined as an utterance, since TED talk usually has a single speaker. If the utterance is too short (2 seconds or less), we concatenate adjacent utterances into one. For sitcoms, we define consecutive sentences from the same speaker as an utterance.


\paragraph{Visual cues}
We compose visual cues with facial expressions and scene descriptions to perceive human-specific and scene-wide contextual information.
Specifically, to process human-specific information, we utilize the active speaker detection algorithm~\cite{tao2021someone} and face detector~\cite{zhang2017s3fd} to crop the face of the speaking person in each video segment. This process effectively identifies the active speaker, especially for sitcoms where many people appear in a single scene, allowing to align visual features with utterances.\footnote{We provide these face-cropped video segments in our dataset.}
For facial expression description, we extract 14 facial action units (FAUs)~\cite{yao2021action}\footnote{We use \href{https://github.com/CVI-SZU/ME-GraphAU}{https://github.com/CVI-SZU/ME-GraphAU} to extract FAUs.} 
from each frame in the video segment with 10 frames per second (FPS).

Then, we accumulate them and take the three most dominant units. 
For scene-wide contextual cues, we use the video captioning 
\cite{wang2022internvideo} to extract scene description. The scene description provides high-level context for the visual cues including 
the surrounding objects and background that interact with the speaker.

\paragraph{Acoustic cues}
We extract the mean and the variance of pitch, intensity, jitter and shimmer as acoustic features from speech utterance using off-the-shelf speech processing models~\cite{arias2017parkinson, dehak2007modeling}. Since the extracted values are real numbers, we initially try to convert them to a linguistic format with certain criteria (\textit{e.g.}, map to "high pitch" if the mean pitch value is greater than 200). However, it is challenging to set an objective criterion that considers various factors, including the speaker's gender, context, and identity. Instead of putting 
real numbers into text, we use themselves as acoustic features by giving a description of them as a prompt to LLMs, leveraging their knowledge on understanding numerical number~\cite{brown2020language, liu2023pre, jiang-etal-2020-know, wallace-etal-2019-universal} (See bold text in parentheses on the $t$ in Figure~\ref{fig:task_definition}).

\section{Annotation for Laughter Reason} \label{app:B}
\begin{figure}
    \centering
    \includegraphics[width=0.2\textwidth]{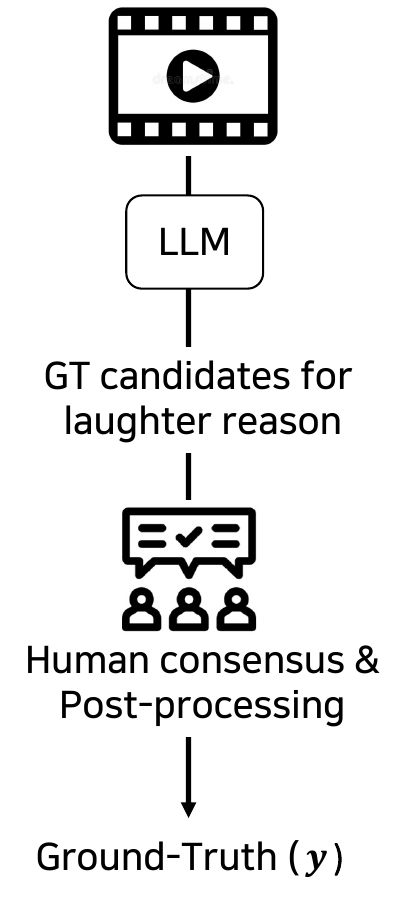}
    \caption{\textbf{Annotation pipeline for laughter reason.}}
    \label{fig:app_ann_pipe}
\end{figure}
We elaborate
the procedure for obtaining
laughter reason consensus (ground-truth; GT) by utilizing large language models’ general knowledge and incorporating it into human consensus. This procedure consists of three steps: (1) build GT candidates, (2) human annotation, and (3) post-processing (See Figure~\ref{fig:app_ann_pipe}).  


\begin{figure*}[t]
    \centering
    \includegraphics[width=1.0\textwidth]{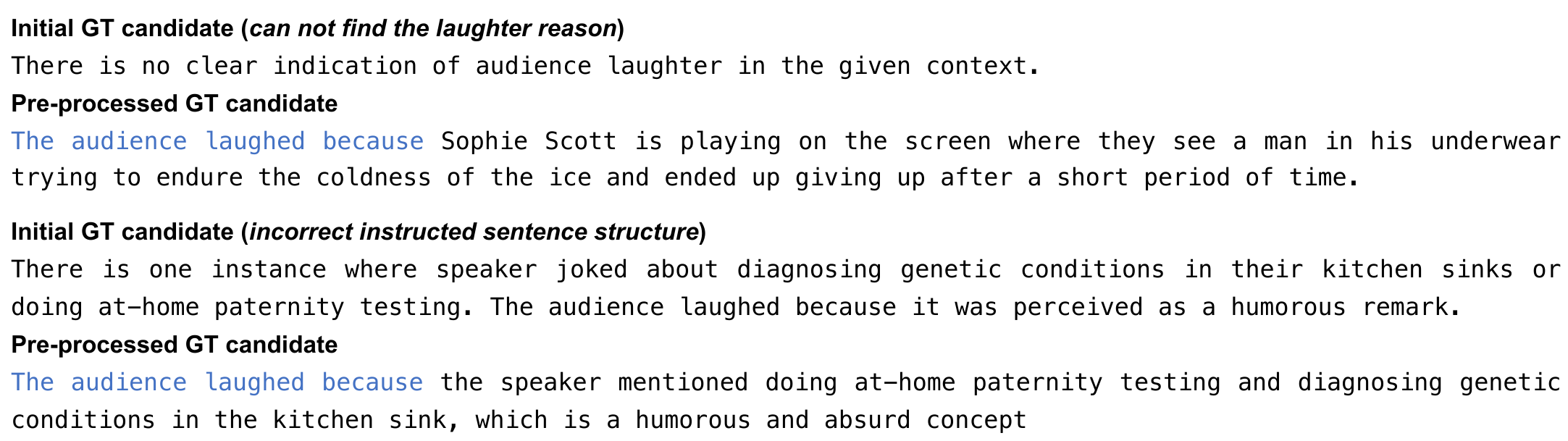}
    \vspace{-6mm}
    \caption{\textbf{Examples of the pre-processing of GT candidates.}  
    In the example above, GPT3.5 fails to infer the reason for the laughter given a multimodal textual representation of the video clip. We handle this by utilizing GPT4 to generate reasons for laughter from the same input. In the example below, the sentence structure does not start with ``The audience laughed because'', which is the structure we want. In this case, we manually revise it for consistent sentence structure.
    }
    \label{supp_fig:processed}
    \vspace{-3mm}
\end{figure*}
\begin{figure*}
    \centering
    \includegraphics[width=1.0\textwidth]{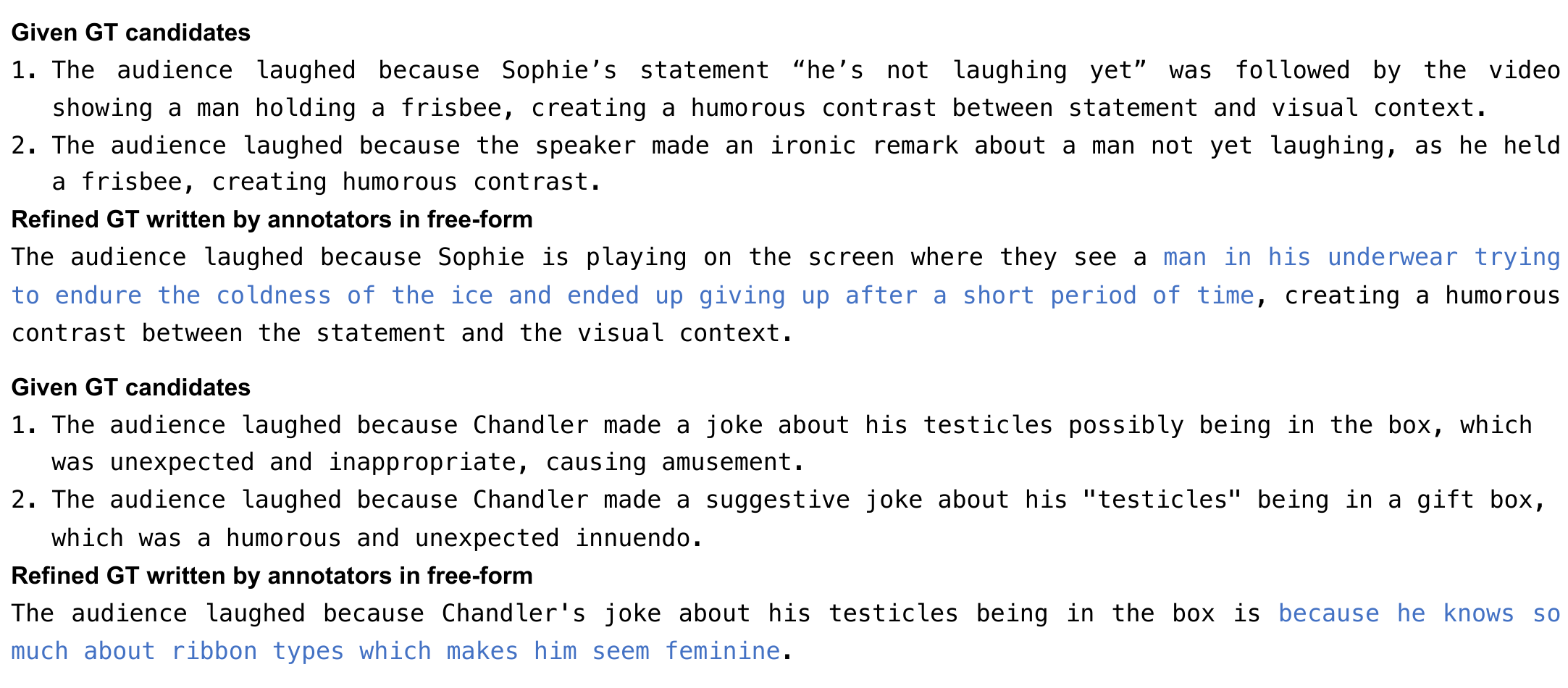}
    \vspace{-7mm}
    \caption{\textbf{Examples of the correction of laughter reason by annotators.} All given GT candidates are passed to the annotators after pre-processing. The free-form responses capture additional visual details (above) and provide a context of why saying ``testicles'' evokes laughter (below).}
    \label{supp_fig:free_form}
    \vspace{-3mm}
\end{figure*}
\begin{figure*}[t]
    \centering
    \includegraphics[width=1.0\textwidth]{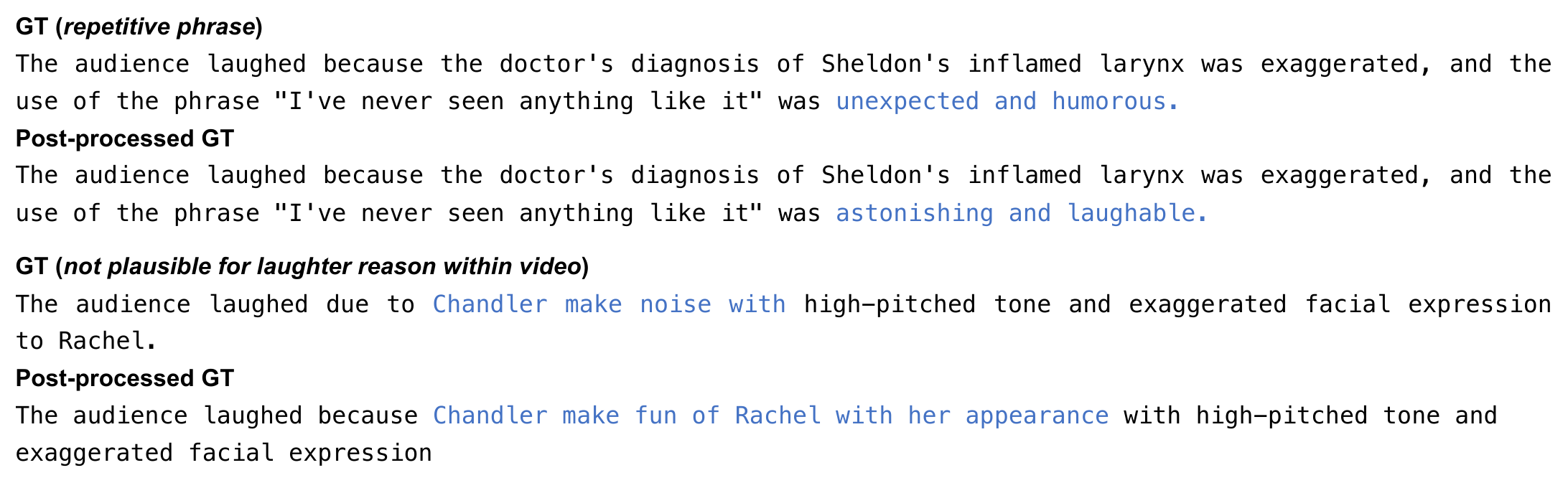}
    \vspace{-7mm}
    \caption{\textbf{Examples of the post-processing on GT.} The example above shows replacing a repetitive phrase with a synonym. The example below shows how we rectify GT when the reason for laughter does not align with the video context.} 
    \label{supp_fig:post}
    \vspace{-3mm}
\end{figure*}
For (1) building GT consensuses, we utilize the large language model (GPT-3.5~\cite{InstructGPT}) with multimodal textual representation $t$ to generate two candidates for the laughter reason. We manually pre-process these candidates if they are invalid or have incorrect sentence structure (See Figure~\ref{supp_fig:processed}). 

For (2) human annotation, 
the processed GT candidates are subsequently presented to annotators from Amazon mechanical turk (AMT) with the corresponding video clip. The annotators are asked to choose the most appropriate explanation among them. 
If the annotators judge that no candidates are appropriate, we instruct the annotators to write or refine the reason in free form. After annotation, the candidate with the most votes is selected as the GT. If at least one annotator provided the reason for laughter in free-form, we manually checked their validity and reflected
them into 
GT. 
Figure~\ref{supp_fig:free_form} shows that free-form responses capture additional visual details and provide an understanding of why certain words 
elicit laughter. See Appendix~\ref{app:F_AMT} for details about AMT.




For (3) post-processing, we additionally verify all GTs for laughter reasons and manually refine it if it is not plausible for laughter reasons with video or has repetitive phrases that might induce spurious correlation. To mitigate this, we replace repeated phrases with synonyms, which are randomized among multiple synonyms. For example, one of the repetitive phrases ``unexpected and humorous'', is randomly replaced with synonyms such as ``astonishing and laughable'', or ``hilarious''. 
As another correction,
even with the best efforts of human annotators, some reasons are not perfectly matched with the video. Figure~\ref{supp_fig:post} shows the post-processing that corrects these kinds of errors.


\paragraph{Annotation quality control}
We use qualification criteria to ensure the annotation quality. We allow annotators from (AU, CA, NZ, GB, US), which represent the English-speaking countries.\footnote{This is because all the video clips in our dataset are in English.}
Additionally, we only allow experienced annotators who are with 10K approved previous HITs and a minimum acceptance rate of 97\% on their previous HITs.
We pay each annotator 0.3 USD(\$) per accepted HIT.







\section{Data Analysis} \label{app:C_data_analysis}
We further conduct a human evaluation to understand our dataset better. Given the video clip, the annotators are requested to determine the laugh. The laugh type annotation explains the distinct characteristics of laughter in TED and sitcoms.

We consider two laugh types: 1) \emph{Release-Triggered Laughter}~\cite{freud1960jokes, fry2011sweet, mindess2017laughter} that results from the alleviating tension amidst constraints such as awkward or complex situation and 2) \emph{Hostility-Triggered Laughter}~\cite{gruner1978understanding, billig2005laughter} that arises from claiming superiority over someone or something, based on ``great families'' of theories of humor~\cite{attardo2008primer}, and ask annotator to determine which one is more appropriate for laughter in video.\footnote{During annotation, we provided full descriptions of the concepts of the laughter types, rather than using the terms.}

\begin{figure}
    \centering
    \includegraphics[width=0.4\textwidth]{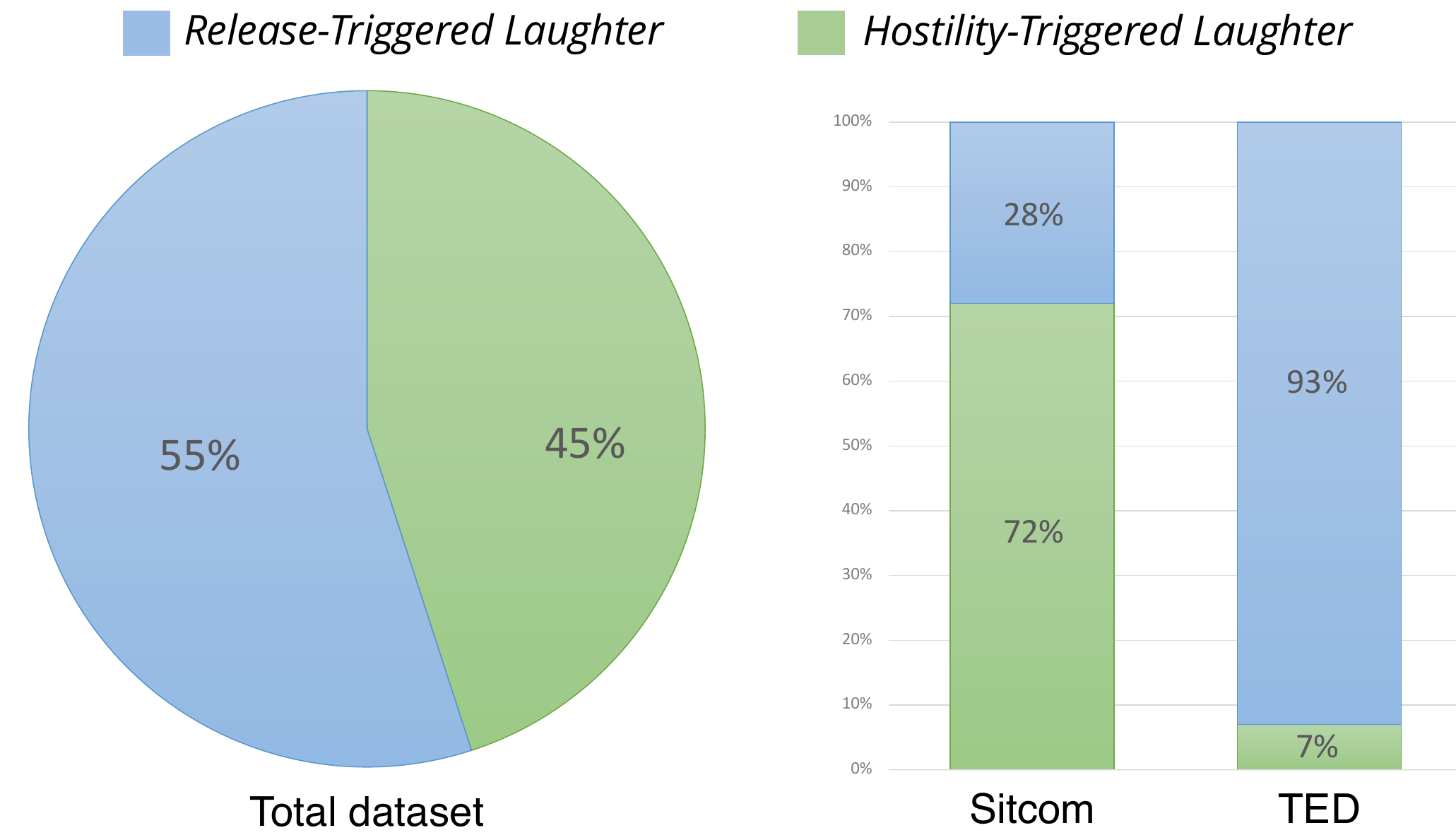}
    \caption{\textbf{Laughter types in our dataset.} Sitcoms tend to have more hostility-triggered laughter, while TED talks have more released-triggered laughter.}
    \label{fig:data_analysis_laugh_type}
    \vspace{-5mm}
\end{figure}

Statistics in Figure~\ref{fig:data_analysis_laugh_type} suggest that sitcoms and TED talks are dominated by different types of laughter, suggesting that the nature of laughter varies by video type. Specifically, the major laugh type in sitcoms is closer to the hostility-induced laughter, and we postulate that sitcoms are typically designed to be entertaining, focusing on humorous situations, witty dialogue, and comedic conflicts among characters. On the other hand, TED talks are dominated by release-triggered laughter. We hypothesize that the talks aim to captivate and engage the audience by releasing constraints and unexpected revelations, creating a dynamic and thought-provoking experience. This type of humor helps maintain interest, and breaks the monotony~\cite{wanzer2010explanation}. By merging these two heterogeneous video types, we can cover a wider range of reasons behind the audience's laughter.

\section{Implementation details} \label{app:D_implementation}

\begin{figure*}
    \centering
    \includegraphics[width=1.0\textwidth]{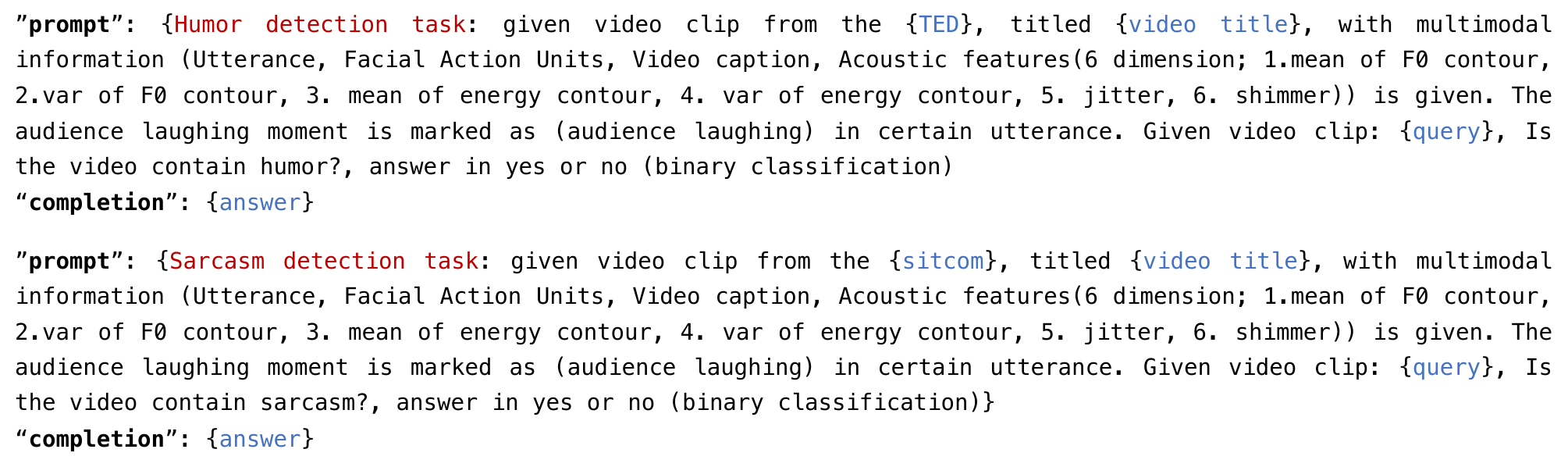}
    \caption{\textbf{Prompt for humor and sarcasm detection.} We manually change video types (sitcom or TED) and video title (such as Patrick Chappatte (2010 Global) or BBT) using the meta information of video clips. The query stands for multimodal textual representation $m$ of the video clip. Answer denote label ($yes$ or $no$) from UR-FUNNY~\cite{hasan-etal-2019-ur} and MUStARD dataset~\cite{castro-etal-2019-towards}.}
    \label{supp_fig:prompt_detection}
    \vspace{-4mm}
\end{figure*}

\begin{table*}[t]
  
  \centering
\vspace{2mm}
  
  \resizebox{0.95\textwidth}{!}{\begin{tabular}{llcccccc}
    \toprule
    Test dataset & Train dataset & Modality & BLEU$_4$ ($\uparrow$) & METEOR ($\uparrow$) & ROUGE$_L$ ($\uparrow$) & BERTScore (F1) ($\uparrow$)\\
    \midrule						
    \multirow{4}{*}{SMILE$_{\text{Sitcom}}$} & \multirow{2}{*}{SMILE$_{\text{Sitcom}}$} & T  & 0.214 & 0.248 & 0.429 & 0.489\\
     && A+V+T  & 0.290 & 0.288 &0.485 &0.548\\
    & \multirow{2}{*}{SMILE} & T  & 0.241 & 0.252 & 0.446 & 0.510\\
     && A+V+T  & \textbf{0.298} & \textbf{0.289} & \textbf{0.499 }&\textbf{0.555}\\
    \cmidrule{1-7}
       \multirow{4}{*}{SMILE$_{\text{TED}}$} & \multirow{2}{*}{SMILE$_{\text{TED}}$} & T & 0.260 & 0.241 & 0.432 &0.459 \\
     && A+V+T  &\textbf{ 0.279} & \textbf{0.260} & \textbf{0.454} & 0.457\\
    & \multirow{2}{*}{SMILE} & T  &0.249 &0.245 & 0.423 & 0.454\\
     && A+V+T  & 0.273 & 0.247& 0.438 &\textbf{0.468}\\
    \bottomrule
  \end{tabular}}\\
    \vspace{1mm}
    \small (a) Video type-wise evaluation \\
    \vspace{2mm}

  \resizebox{0.95\textwidth}{!}{\begin{tabular}{llcccccc}
    \toprule
    Test dataset & Train dataset & Modality & BLEU$_4$ ($\uparrow$) & METEOR ($\uparrow$) & ROUGE$_L$ ($\uparrow$) & BERTScore (F1) ($\uparrow$) \\
    \midrule						
    \multirow{2}{*}{SMILE$_{\text{Sitcom}}$} & SMILE$_{\text{TED}}$ & A+V+T  & 0.161 & 0.254 & 0.390 & 0.407\\
     & SMILE$_{\text{Sitcom}}$ & A+V+T & 0.290 & 0.288 &0.485 &0.548\\
   \cmidrule{1-7}
\multirow{2}{*}{SMILE$_{\text{TED}}$} & SMILE$_{\text{Sitcom}}$ & A+V+T  & 0.153 & 0.193 & 0.369 & 0.449 \\
     & SMILE$_{\text{TED}}$ & A+V+T  & 0.279 & 0.260 & 0.454& 0.457\\

    \bottomrule
  \end{tabular}}\\
  \vspace{1mm}
    \small (B) Cross-dataset evaluation\\

  \caption{\textbf{Analysis on video types.}  
  In (a), we conduct the video type-wise evaluation as the dominant laughter type differs along the video type. In (b), we evaluate the model by testing on the different video types, \ie, cross-dataset.}    
  \label{tab:analysis}
\end{table*}

\begin{table}
\centering
\resizebox{1.0\linewidth}{!}{
  \begin{tabular}{lll|cc} 
    \toprule
    Test & A & B & A wins (\%) & Fleiss'-$\kappa$ \\
    \cmidrule{1-5}
    TED & GPT-3 (SMILE) & GPT-3 (TED)  & 66.2 & 0.40\\
    Sitcom & GPT-3 (SMILE) & GPT-3 (sitcom)  & 61.4 & 0.63 \\
    \bottomrule
  \end{tabular}
}
\caption{\textbf{Pairwise human evaluation.} We compare the model trained with the whole dataset (SMILE) with a subset (TED, sitcom) and evaluate them with the test set of each subset.}
\label{App:tab:human_winrate}  \vspace{-4mm}
\end{table}
\paragraph{GPT3 fine-tuning}
We utilize the OpenAI fine-tuning API and fine-tune \emph{davinci}. The prompt for fine-tuning is the same as the aforementioned experiments. 
We follow the fine-tuning scheme provided on the OpenAI webpage.\footnote{\href{https://platform.openai.com/docs/guides/fine-tuning}{https://platform.openai.com/docs/guides/fine-tuning}; OpenAI has not opened the details of the API's fine-tuning mechanisms, which is currently hidden. 
}

\paragraph{LLaMA fine-tuning}
LLaMA is LLM, an open-source model for research.  
We fine-tune the full parameters of LLaMA for 5 epochs. We utilize 4 A100 (80GB) for distributed fine-tuning with batch size 4 per device and a learning rate 1e-4. We also leverage fp16 mixed precision.

\paragraph{Video-LLaMA fine-tuning}
We use Video-LLaMA which consists of pre-trained Blip2, Vicuna-13B, and Imagebind-huge. We train audio, video Q-former, and projection layers while other parameters are frozen. We utilize 8 A100 (80GB) for distributed fine-tuning with batch size 1 per device and an initial learning rate (3e-5), and weight decay (0.05) for 10 epochs. We also leverage mixed precision that uses fp16 for multiplication and fp32 for addition.

\paragraph{Detection}
For the sarcasm~\cite{castro-etal-2019-towards} and humor detection~\cite{hasan-etal-2019-ur} tasks, 
we finetune LLaMA-13B~\cite{llama} and GPT-3~\cite{GPT3} with our multimodal textual representation. 
GPT-3 finetuning is as same as described for the laugh reasoning task.
For LLaMA-13B, we follow the fine-tuning script on Vicuna~\cite{vicuna2023}\footnote{\href{https://github.com/lm-sys/FastChat}{https://github.com/lm-sys/FastChat}}. Examples of the prompts for both tasks that cast classification task to generation task are shown in Figure~\ref{supp_fig:prompt_detection}. We use four A100 (80GB) for each training. 
We follow Vicuna's default LLaMA fine-tuning hyperparameters except for setting the per-device batch size to 3 and the number of training epochs to 20.
\begin{figure*}[t]
    \centering
    \includegraphics[width=\textwidth]{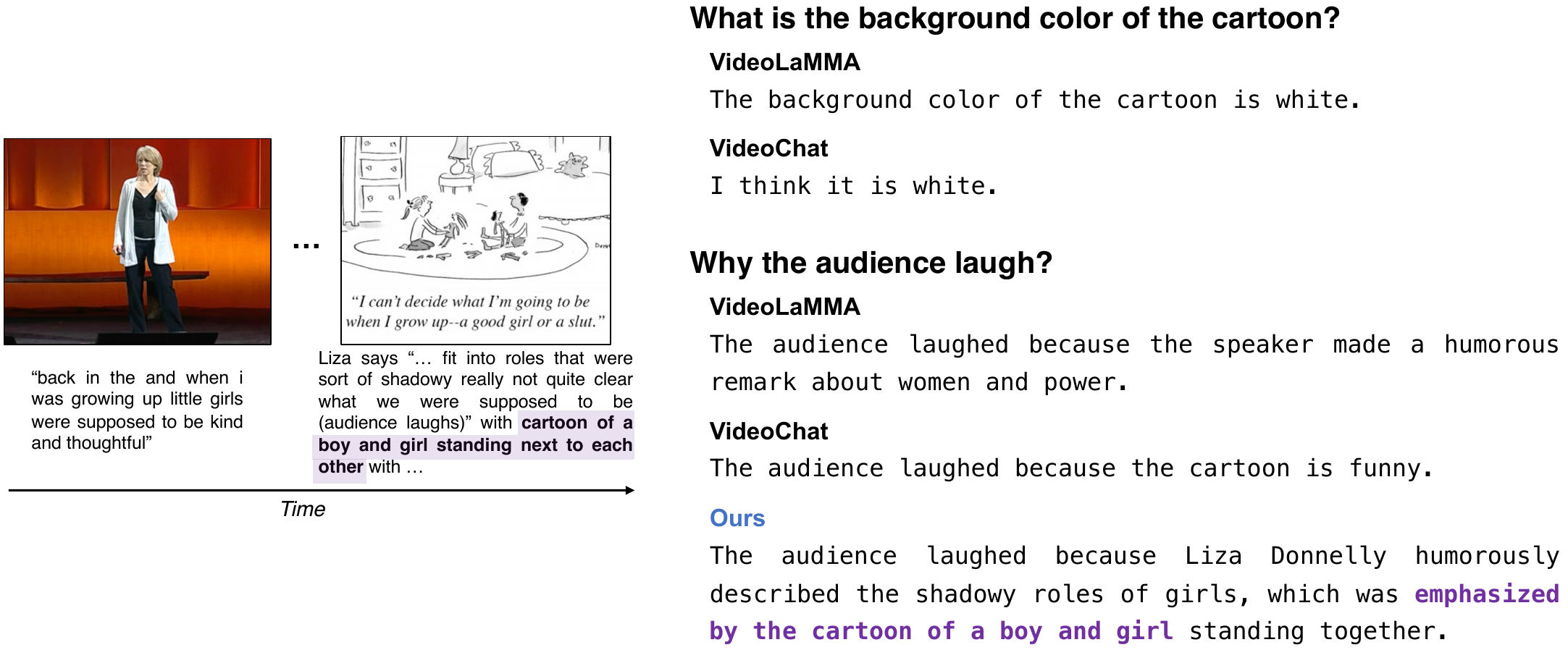}
    \caption{\textbf{Examples of laugh reasoning on recent video language models.} 
    While recent video language models, such as Video-LLaMA~\cite{damonlpsg2023videollama} and VideoChat~\cite{2023videochat}, can respond to the general question in the video, they struggle to plausibly explain the reason for the laughter in the video.}
    \label{fig:supqual}
\end{figure*}

\section{Additional Experiments} \label{app:E_addi_exp}

\paragraph{Evaluation by video types}
The type of laughter varies depending on the source of the video, as shown in Figure~\ref{fig:data_analysis_laugh_type}. To explore this further, we evaluate each video type independently. Instead of fine-tuning GPT3 on the entire SMILE dataset, we separately fine-tune the models on subsets of the dataset, namely SMILE$_{\text{Sitcom}}$ and SMILE$_{\text{TED}}$. As summarized in \Tref{tab:analysis}~(a), even when models are independently fine-tuned to different video types, their performance is comparable to that of the model trained on the SMILE dataset. Interestingly, in the human evaluation, the model trained on whole data (SMILE) is preferred over the model trained on each video type. This suggests that our dataset, SMILE, covers the diverse laughing characteristics to lead GPT3 to learn generalized laughter reasons across different types of videos.

However, we observe that testing the model across different video types, \eg, training on SMILE$_{\text{Sitcom}}$ and testing on SMILE$_{\text{TED}}$, results in a significant performance drop, as shown in \Tref{tab:analysis}~(b). We speculate that this is due to differences in laughter types presented in each source video. This supports the idea that combining these two heterogeneous video types could help the model learn to understand a broader range of reasons behind audience laughter.


\paragraph{Video language model}
While the previous methods~\cite{zellers2019recognition,zadeh2019social} have aimed to learn and reason about social interactions from visual data, they formulate the task in multiple-choice setups. 
By virtue of the advance of large language models, 
recent work has suggested multimodal models capable of generating natural language responses to questions about a video,
rather than outputting a multiple-choice answer.
In this context, we examine if these models can exhibit the capability to reason behind laughter in a given video. 
We feed the same video from Figure~\ref{fig:qual} into recent video-language (VL) models, Video-LLaMA~\cite{damonlpsg2023videollama}\footnote{\href{https://github.com/DAMO-NLP-SG/Video-LLaMA}{https://github.com/DAMO-NLP-SG/Video-LLaMA}} and VideoChat~\cite{2023videochat}\footnote{\href{https://github.com/OpenGVLab/Ask-Anything}{https://github.com/OpenGVLab/Ask-Anything}}, and showcase their generated reasoning in Figure~\ref{fig:supqual}. While these models can respond to general questions about the video, they struggle to reason about moments of laughter. Unlike existing multimodal reasoning work, we contribute a new perspective to multimodal reasoning, aiming to understand and reason about an important social signal, laughter.

\section{Human annotation from Amazon Mechanical Turk} \label{app:F_AMT}
\begin{figure*}
    \centering
    \includegraphics[width=1.0\textwidth]{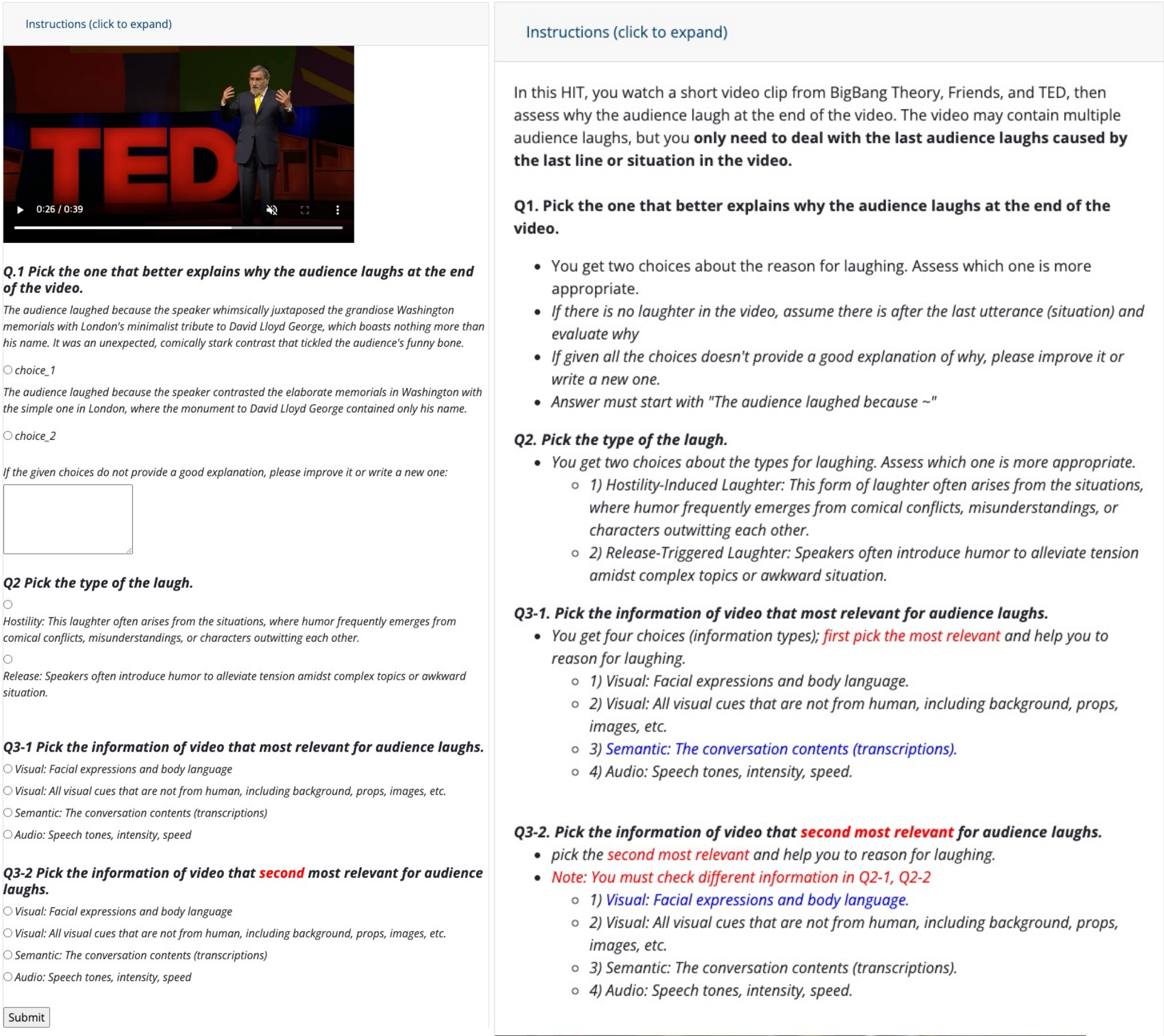}
    \caption{\textbf{Examples of the AMT interface (left) and instructions (right) that the annotators worked on.} The annotators are asked to watch the video clip and answer the three questions. The third question is split into two parts. We put the instructions at the top of the interface to emphasize how the annotators should answer each question.}
    \label{supp_fig:AMT}
\end{figure*}

Figure~\ref{supp_fig:AMT} shows our interface and instructions for the annotators working on Amazon Mechanical Turk (AMT).
We define a questionnaire per video clip as a Human Intelligence Task (HIT).
We ask AMT annotators three questions in a HIT, 1) laughter reason, 2) laugh type, and 3) the multimodal cues in perspective of which cues are related to laughter in the video. 
The first question is for obtaining GT annotations for laughter reasons and pairwise human evaluation in \S~\ref{sec:exp}.
The second and third questions are for the data analysis purpose, which provides further understanding of our dataset (See \S~\ref{sec34} in the main paper and Appendix~\ref{app:C_data_analysis}).

\end{document}